\documentclass{article}

\usepackage[draft]{todonotes}   
\usepackage{bm}
\usepackage{mathtools}
\usepackage{microtype}
\usepackage{graphicx}
\usepackage{subfigure}
\usepackage{booktabs} 
\usepackage{amssymb} 
\usepackage{amsmath}
\usepackage{amsthm}
\DeclareMathOperator*{\argmax}{arg\,max}
\DeclareMathOperator*{\argmin}{arg\,min}
\usepackage{hyperref}
\usepackage{bbm}
\newtheorem{proposition}{Proposition}[section]
\newtheorem{theorem}[proposition]{Theorem}

\newcommand{\defeq}{\vcentcolon=}

\usepackage[accepted]{icml2018}

\icmltitlerunning{Welfare and Distributional Impacts of Fair Classification}

\begin{document}

\twocolumn[
\icmltitle{Welfare and Distributional Impacts of Fair Classification}



\icmlsetsymbol{equal}{*}

\begin{icmlauthorlist}
\icmlauthor{Lily Hu}{har}
\icmlauthor{Yiling Chen}{har}
\end{icmlauthorlist}

\icmlaffiliation{har}{School of Engineering and Applied Sciences, Harvard University, Cambridge, MA, USA}

\icmlcorrespondingauthor{Lily Hu}{lilyhu@g.harvard.edu}

\icmlkeywords{Machine Learning, ICML}

\vskip 0.3in
]



\printAffiliationsAndNotice{}  

\begin{abstract}
Current methodologies in machine learning analyze the effects of various statistical parity notions of fairness primarily in light of their impacts on predictive accuracy and vendor utility loss. In this paper, we propose a new framework for interpreting the effects of fairness criteria by converting the constrained loss minimization problem into a social welfare maximization problem. This translation moves a classifier and its output into utility space where individuals, groups, and society at-large experience different welfare changes due to classification assignments. Under this characterization, predictions and fairness constraints are seen as shaping societal welfare and distribution and revealing individuals' implied welfare weights in society---weights that may then be interpreted through a fairness lens. The social welfare formulation of the fairness problem brings to the fore concerns of distributive justice that have always had a central albeit more implicit role in  standard algorithmic fairness approaches.
\end{abstract}

\section{Introduction}
\label{intro}
In his 1979 Tanner Lectures, Amartya Sen noted that since nearly all theories of fairness are founded on an equality of \textit{some} sort, the heart of the issue rests on clarifying the ``equality of what?'' problem \yrcite{sen}. The field of fair machine learning has not escaped this essential question. Do such tools have an obligation to assure probabilistic equality of outcomes \cite{feldman2015certifying,hardt2016equality}? Or do they simply owe an equality of treatment \cite{dwork2012fairness}? Does fairness demand that individuals (or groups) be subject to equal mistreatment rates \cite{zafar2017fairness,bechavod2017learning}? Or does being fair refer only to avoiding some intolerable level of discrimination? Differential demands of fairness contrast starkly with each other in both their effects on the outcomes that are ultimately issued and in the means by which they may be implemented. 

In machine learning, the task of accounting for fairness involves comparing myriad metrics---probability distributions, error likelihoods, classification rates---sliced every way possible to reveal the range of inequalities that may arise before, during, and after the learning process. But as shown in Chouldechova \yrcite{chouldechova2017fair} and Kleinberg et al. \yrcite{kleinberg2017inherent}, fundamental statistical incompatibilities rule out any solution that satisfies all parity metrics, and we are left with the harsh but unavoidable task of adjudicating between these measures and methods. Past work has been limitedly able to address these ``inherent trade-offs.'' For one, the leading approach of constrained loss minimization offers little guidance by itself for choosing among the fairness desiderata, which appear incommensurable and result in different impacts on different individuals and groups. Even when comparisons are made across fairness metrics and methods, current approaches refer only to losses in predictive accuracy or vendor utility as illustrative of the cost of assuring different types of fairness. This methodology flattens multifaceted distributive procedures, which involve many individuals and thus many interests, into a two-dimensional comparison of accuracy vs. ``fairness,'' and as a result fails to capture the fundamentally social nature of fairness. 

This paper proposes a conceptual framework and methodology for conceiving of fairness in machine learning that is based in analysis of the society-wide distributional effects of classifier outputs. Our approach maps the standard empirical risk minimization task in supervised learning into a corresponding social welfare maximization task as it appears in welfare economics' Planner's Problem. Social welfare functionals are typically formulated as the sum of weighted utility functions, where an individual's weight represents the value placed by society on her welfare. Inverting the Planner's Problem of efficient social welfare maximization generates a question that is more concerned with social equity: \textit{``Given a particular allocation, what is the presumptive social weight function that would yield the allocation as optimal?''} Mapping a fair learning problem into a social welfare problem lends new insight into the different fairness regimes proposed by admitting comparison of the distributive and welfare effects of the various models and constraints used in prediction. By centering social welfare as the primary object of interest of fair machine learning, we highlight a positive conception of fairness as a societal good rather than as an oppositional force that detracts from a decision-maker's accuracy and optimality. We believe that this perspective presents a more nuanced and faithful understanding of fairness as a social ideal.

In this paper, we establish this mapping for the task of binary classification using linear SVMs. Starting with standard empirical risk minimization, we present general characteristics of the structure of the implied welfare weight function corresponding to a given learned classifier. We then follow the two main distinct approaches to fairness-adjusted classification, connecting the altered outcomes and margins resulting from each of these new methodologies with shifting weight functions in the social welfare problem. In offering two different perspectives on how the welfare weight distribution may be transformed by fair adjustment, we present novel interpretations of how fairness constraints alter boundary-based classifiers' treatments of individuals, groups, and the underlying feature space.

The deployment of socially-oriented machine learning inevitably implicates several ethical questions surrounding the tension between shared societal norms and ideals and a decision-maker's private goals and interests. Most leading methodologies have focused on optimization of utility or welfare to the vendor, limiting our ability to answer questions about how individuals, groups, and society-at-large fare under various distributive allocations. The social welfare perspective directly engages both questions of efficiency, in the task of maximization, and equity, in the design of welfare weights. This perspective is especially enlightening when applied to sectors in which the government, acting as the Planner, maintains a strong interest in issues of distributive fairness and can justifiably make interpersonal comparisons of utility. Financial services, wherein loan and credit approvals are increasingly automated, satisfy both criteria and will be the main application focus of this paper.



\section{Problem Formalization}
\label{mapping}
Before we formalize this paper's objective of connecting loss minimization with social welfare maximization, we provide an overview of the separate perspectives on and methodologies for achieving fairness in optimal predictions or allocations. In this paper, we will center our analysis on binary decision tasks using linear SVM classifiers. 

Consider risk minimization within the supervised learning classification setting where the decision-maker seeks the classifier that minimizes the probability of error on a training set of $n$ data points $\mathcal{D} = \{\mathbf{x}_i, y_i\}_i^n$. The risk-minimizing predictor is thus $\hat{h_{\bm{\theta}}}\defeq \argmin_{h\in \mathcal{H}} \sum_{i=1}^n \ell(h(\mathbf{x}_i), y_i)$ where $\mathcal{H}$ contains only those classifiers that are linear halfspaces that may be written as $h_{\bm{\theta}}(\mathbf{x}) = {\bm{\theta}}^\intercal \mathbf{x} + b$ with $\mathbf{x}, \bm{\theta} \in \mathbb{R}^d$ and $b \in \mathbb{R}$. For binary classification, the ultimate classifications follow $\hat{y}^{ML}(\mathbf{x}) = sgn(\hat{h_{\bm{\theta}}}(\mathbf{x}))$. For our considered case of linear SVMs, we will relax $0-1$ loss and replace it with hinge loss $\ell_h = \max (0, 1-y_ih_{\bm{\theta}}(\mathbf{x}_i))$.


In the social welfare problem, a Planner is charged with the task of maximizing societal welfare given as an aggregate weighted sum of individuals' utilities, $W = \sum_{i=1}^n w_i u_i$. The Planner distributing financial loans solves ${y}^{SWF}(\mathbf{x}) \defeq \argmax_{y_i} \sum_{i=1}^n w_iu(x^m_i, y_i)$, where individual $i$'s contribution to society's overall welfare is a product of her utility $u(x^m_i, y_i)$, a function of her income and allocation outcome, and her societal weight $w_i$. In binary classification, the Planner can either allocate the good to individual $i$ or not ($y_i \in \{0,1\}$), while in the standard welfare problem, the Planner faces a fixed exogenous budget for allocations. In our formulation, the budget is set to be equal to the number of positive instances issued by the classifier. 

Our central question of interest in thus: For a given boundary classifier ${h}_{\bm{\theta}}$ output by a loss minimization task and a set of income levels $x^m_i$, can we characterize aspects of the functional form of the welfare weights $w_i$ within $W = \sum_{i=1}^n w_i u(x^m_i, y_i)$ that would yield an optimal social allocation such that $\hat{y}^{SWF} = \hat{y}^{ML}$? We call such an allocation produced by a learned classifier that is also socially optimal in the welfare sense a \textit{matched allocation}.

\subsection{Preliminary Results}
A mapping from loss minimization to social welfare maximization requires that welfare weights be formulated to ``track" the classifier's treatment of individuals. The Planner with a given classifier $h(\mathbf{x})$ must prefer individual $i$ to $j$ whenever $h(\mathbf{x}_i) > h(\mathbf{x}_j)$; under matched allocations, equivalent margins enforce equivalent weights. The Planner considers both classification margins $h(\mathbf{x}_i)$ and incomes $x^m_i$, and as such, weights must be a function of both. Formally, the marginal social gain associated with a positive classification $y = 1$, $w_f(h(\mathbf{x}),x^m) = w(h(\mathbf{x}), x^m)\Delta u (x^m)$, where $\Delta u(x^m) = u(x^m, 1) - u(x^m, 0)$ represents the utility gain due to receiving a loan, satisfies
\begin{equation}
\label{eq:dwdh}
dw_f= \Delta u[\frac{\partial w}{\partial h}dh+ \frac{\partial w}{\partial x^m} dx^m] + w\frac{d\Delta u}{d x^m}dx^m > 0
\end{equation}
whenever $dh > 0$ and $dw_f = 0 \Leftrightarrow dh =0$. From here, characterizing welfare weights $w$ depends on the  functional form of $u(x^m, y)$.

\section{Results}
\label{results}

\subsection{Unconstrained Loss Minimization}

In the simplest case in which $u(x^m, y)$ is either linear 
or additively separable,
then $\frac{d\Delta u}{d x^m} =0$, and Eq (\ref{eq:dwdh}) reduces to the condition that welfare weights $w = f(h(\mathbf{x}))$ where $f$ is any positive monotonic transformation of $h(\mathbf{x})$. Here, weights do not depend on $x^m$ at all ($\frac{d w}{ d x^m} = 0$), and so long as the welfare weights are such that $\frac{dw_f}{dh} = \frac{d w}{d h}> 0$, the Planner is justified in distributing the matched allocation. 

\subsubsection{$u(x^m, y)$ concave in $x^m$}
When utility exhibits the property of diminishing marginal utility of income such that $u(x^m,y)$ is concave in $x^m$, the implied welfare weight function must now explicitly account for $x^m$. In the binary allocation setting, the standard statement of concavity, $\frac{\partial^2 u}{(\partial x^m)^2} < 0$, becomes equivalent to $\frac{d \Delta u}{d x^m} < 0$. Under this assumption of $u$, the welfare weight condition in Eq. (\ref{eq:dwdh}) may be expanded to
\begin{align}
\label{eq:concave}
\frac{\partial w}{\partial h}dh + \frac{\partial w}{\partial x^m} dx^m> |\frac{w}{\Delta u}\frac{d \Delta u}{d x^m} dx^m| 
\\
\label{eq:concave2}
\frac{\partial w}{\partial x^m} dx^m = |\frac{w}{\Delta u}\frac{d \Delta u}{d x^m} dx^m| 
\end{align}
such that $dh>0$ enforces a strictly greater lower bound on $dw_f$ compared to the linear case, and changes in classification margin $h$ must correspond to larger deviations in $w$. Notice that when utility is linear or additively separable, two individuals $i$ and $j$ with identical classification margins would be equally preferred under welfare weights $w(h(\mathbf{x}_i), x^m_i) = w(h(\mathbf{x}_j), x^m_j)$ and would receive the matched allocation under the Planner's Problem even if they were endowed with differing income levels $x^m_{i}>x^m_{j}$. In contrast, under concave $u(x^m, y)$, the condition that $\frac{\partial w}{\partial h} dh > 0$ is insufficient to achieve the appropriate welfare weights. $\frac{\partial w}{\partial x^m}$ must also be increasing in the the concavity of $u(x^m, y)$ as given in Eq. (\ref{eq:concave2}). As marginal utility returns to income decrease, the Planner is only justified in allocating the loan to an individual with high $x^m$ if she inflates the individual's welfare weight in accordance with her income to ``offset'' the loss due to concavity. These relations and conditions of the weight function $w(h(\mathbf{x}), x^m)$ in the Planner's Problem are summarized in the following Theorem.
\begin{theorem}
\label{theorem1}
Given classifier margins $h(\mathbf{x}_i)$ and income levels $x_i^m$, the following welfare maximization problem
\begin{displaymath}
    \begin{aligned}
        {y}^{SWF} = \argmax_{{y}_i} &\sum_{i=1}^n w(h(\mathbf{x}_i), x^m_i)u(x^m_i, {y_i})\\
        \text{s.t} &\; \sum_{i=1}^n \mathbbm{1}_{\{y^{SWF}_i>0\}} = \sum_{i=1}^n \mathbbm{1}_{\{h(\mathbf{x}_i)>0\}},\\
                   &\; {y_i} \in \{0,1\}\text{ }\forall i\in [n]
    \end{aligned}
\end{displaymath}
with weights $w(h(\mathbf{x}_i), x^m_i)$ satisfying Eq. (\ref{eq:concave}) and (\ref{eq:concave2})
yields the matched optimal allocation ${y}^{SWF} = {y}^{ML}$. Moreover, the welfare weight function is of multiplicative form $w(h(\mathbf{x}), x^m) = f(h(\mathbf{x}))g(x^m)$ such that $g(x^m)= \frac{k}{{\Delta u}(x^m) }$ for some constant $k \in \mathbb{R}_{+}$.
\end{theorem}
\begin{proof}

Notice that the Planner can reduce her task to a binary knapsack problem (BKP) by considering the maximization $\sum_{i=1}^n w(h(\mathbf{x}_i), x^m_i) v(x^m_i, y_i)$ where 
\[v(x^m_i, {y_i})=
\begin{cases}
0, \text{ if } {y_i} = 0\\
\Delta u(x^m_i), \text{ if } {y_i} = 1
\end{cases}\]
When $u(x^m,y)$ is concave in $x^m$, $\frac{d\Delta u}{d x^m} < 0$. The optimal solution to BKP may be attained via the greedy algorithm in which the Planner allocates the good ${y}_i = 1$ in decreasing order starting with the individual $i$ with the highest marginal contribution to social welfare, $w(h(\mathbf{x}_i), x^m_i)\Delta u(x^m_i)$, until she depletes her ``budget" $\sum_{i=1}^n \mathbbm{1}_{\{h(\mathbf{x}_i)>0\}}$. This procedure generates the same ordering on individuals as $h(\mathbf{x})$ whenever $\frac{d(w(h(\mathbf{x}),x^m)\Delta u(x^m))}{dh} > 0$, which is the inequality condition given in Eq. (\ref{eq:concave}). When two individuals share the same $h(\mathbf{x})$, the greedy algorithm for BKP  must also be indifferent such that $\forall x^m_i, x^m_j \ge 0$, marginal gains are equal: $w(h(\mathbf{x}_i), x^m_i)\Delta u(x^m_i) = w(h(\mathbf{x}_j), x^m_j)\Delta u(x^m_j)$, corresponding to Eq. (\ref{eq:concave2}). Notice that generally $w_f(h(\mathbf{x}), x^m)$ cannot be a function of $x^m$ and as a result, the functional form of $w(h(\mathbf{x}), x^m)$ can be decomposed into functions $f(h(\mathbf{x}))$ and $g(x^m) = \frac{k}{\Delta u(x^m)}$ for some constant $k \in \mathbb{R}_+$. 

The proof follows similarly when $u(x^m, y)$ is linear or additively separable, and any weight function $w(\cdot)$ that preserves the ordering given by $h(\mathbf{x})$ yields the same BKP solution. Since weights are defined up to a constant, this result agrees with the multiplicative decomposition of $w(h(\mathbf{x}), x^m)$. 
\end{proof}

The multiplicative form of these underlying social welfare weights highlights two intertwined effects of using boundary-based classifiers in financial distribution decisions. Concavity of utility enforces a term that explicitly incorporates wealth as having a multiplicative impact on welfare weights. Moreover, the wealth effect encoded in $g(x^m)$ is compounded by the classifier score effect in $f(h)$ such that differences in individuals' classification margins also amplify differences in their incomes, and vice versa, in determining an individual's ultimate social weight. In the binary classification task, intensified disparities in welfare weight magnitudes do not affect the Planner's optimal allocation, but such differences do have significant repercussions in more general welfare maximization settings in which the Planner distributes allocations $y^{SWF}_i \in \mathbb{R}_+$.

\subsection{Fairness-constrained Loss Minimization}
Having characterized aspects of the implied social welfare weight functions under standard loss minimization, we now move to ``fair" formulations of learning and ask how popular parity-based constraints on optimization may be translated into social welfare space where they may be interpreted as redistributive mechanisms that act to shift welfare weight among individuals and groups. 

\subsubsection{Fair Post-processing}

A post-processing approach to fairness proposed by Hardt et al. adjusts the  distribution of outcomes by using sensitive attribute information to construct group-specific thresholds for classification \yrcite{hardt2016equality}. This approach grants flexibility to practitioners who can apply the adjustment without needing to access the original dataset or learn a new classifier. 

Without a new classifier or new margins, welfare weights explicitly incorporate group information to handle fairness criteria. The welfare problem adopts the post-processing approach of resolving fairness constraints by transforming the original $h_0(\mathbf{x})$ margins by group-specific threshold factors to achieve various fairness parities. Since $\frac{\partial w}{\partial h} dh> 0$, then any positive affine transformation $h'(\mathbf{x}, z) = T(h_0(\mathbf{x}); \tau_0, \tau_z)$ applied group-wide, where $\tau_0$ and $\tau_z$ represent the old and new group-specific thresholds respectively, that maps $h_0(\mathbf{x}) = \tau_0$ to new margin $h'(\mathbf{x}, z) = \tau_z$ preserves ordering and interval scales. Then the optimal allocation ${y}^{SWF}$ with weights $w\Big(T(h_0(\mathbf{x}); \tau_0, \tau_z), x^m\Big)$ will match the post-processed fair allocation ${y}^{ML}$. 

\subsubsection{Learning a Fair Classifier} 
Recent efforts have incorporated fairness constraints into the learning process itself by way of regularization \cite{bechavod2017learning} or convex proxies for constrained optimization \cite{zafar2017fairness} to generate a new classifier $h'(\mathbf{x})$. 
We note that conditions for the implied welfare weights under $h'(\mathbf{x})$ may be rederived by direct appeal to Theorem \ref{theorem1}, but we also present in this section an alternative procedure that although requiring the Planner to have access to richer information---she must know both the previous and new classifiers, rather than just individuals' margins---admits derivation of new conditions on $w'(\cdot)$ from old conditions on $w(\cdot)$. This technique allows direct comparison of the two functions $w$ and $w'$ and thus sheds light on how fairness constraints shift the distribution of social welfare weight. 

We derive conditions for $dw_f'(h'(\mathbf{x}), x^m)$ in terms of $dw_f(h(\mathbf{x}), x^m)$ by constructing a transformation from the old margins $h(\mathbf{x})$ to the new margins $h'(\mathbf{x})$. Let $T: h_{\mathbf{x}} \mapsto h'_{\mathbf{x}}$ be the linear transformation that maps the orthogonal projection of $\mathbf{x}$ onto $h$ to its orthogonal projection onto $h'$. Formally, we have $h_\mathbf{x} = P_h \mathbf{x} \in \mathbb{R}^{d}$ where $P_h \in \mathbb{R}^{d\times d}$ gives the orthogonal projection mapping of $\mathbf{x}$ onto $h$. Then following Eq. (\ref{eq:concave}) with $w_f(h'(\mathbf{x}), x^m) = w\Big(D(\mathbf{x}, T(h(\mathbf{x}))), x^m \Big) \Delta u(x^m)$, where $D(\mathbf{x}, T(h_{\mathbf{x}}))$ gives the Euclidean distance from $\mathbf{x}$ to its projection on $h' = T(h_{\mathbf{x}})$, we have that
\begin{equation}
\label{eq:proj}
\frac{\partial w(\cdot) }{\partial D(\cdot)} dD(\cdot) + \frac{\partial w(\cdot)}{\partial x^m} dx^m > | \frac{w(\cdot)}{\Delta u} \frac{d \Delta u}{d x^m}dx^m|
\end{equation}
where $D(\cdot) =  D(\mathbf{x}, T(h_{\mathbf{x}}))$ and $w(\cdot) = w\big(D(\cdot), x^m\big)$. The total differential $dD(\cdot)$ is computed as 
\begin{equation}
\label{eq:dD}
\begin{split}
(\sqrt{\lVert(\mathbf{x}-T(P_h(\mathbf{x})))\rVert^2})^{-1}\Big[\sum_{i=1}^{d}\big[x_i-\sum_{j=1}^{d}T_{ij} \big((P_{h})_{j} \cdot \mathbf{x}\big)\big]  dx_i +
\\
\big[\big(\sum_{j=1}^{d}\big(T_{ij} ((P_{h})_{j} \cdot \mathbf{x}))  - x_i\big)\big(\sum_{k=1}^{d} T_{ik} \big((P_{h})_{k} \cdot d\mathbf{x}\big)\big)\big]\Big]
\end{split}
\end{equation}
Since $P_h$, $P_{h'}$ and $T$ are unique for boundary classifiers $h(\mathbf{x})$ and $h'(\mathbf{x})$, and the Euclidean distance $D(\mathbf{x}, T(h_{\mathbf{x}}))$ is easily computable, all of the new variables and functions in Eq. (\ref{eq:dD}) can be analytically derived. Since $\frac{\partial w(\cdot)}{\partial D(\cdot)} = \frac{\partial w}{\partial h}$, the multiplicative factor given by $dD(\cdot)$ dictates how the new welfare weights $w'(h'(\mathbf{x}), x^m)$ shift. In particular, the inclusion of vector rows $T_{i}$ of the matrix $T$ offers a geometric interpretation of the new classifier's transformation of the feature space. Thus, working from the previous weight function $w(h(\mathbf{x}), x^m)$ corresponding to the original unfair classifier $h$ and ensuring that Eq. (\ref{eq:proj}) and Eq. (\ref{eq:dD}) holds, the new weights $w'(h'(\mathbf{x}), x^m)$ will justify the fair matched allocation $y^{SWF} = y^{ML}$ under social welfare maximization. 

\section{Discussion}
\label{discussion}
Within the financial services sector, both disparate impact and accuracy loss may be understood in terms of utility gains and losses incurred by different agents. By connecting loss minimization of accuracy to social welfare maximization, we make such trade-offs explicit and in doing so, broaden the scope of the fairness question to include distributive justice. This descriptive mathematical link sets the groundwork for the requisite normative reasoning of fair machine learning. Different loss functions, parity constraints, and fairness-adjustment methodologies all differentially impact ``optimal" classifier behavior. Translating these various effects into changes in welfare weights, which may be analyzed at the individual level or summed to reveal group shares of societal welfare, allows practitioners to better interpret and evaluate the distributive impacts of predictions and as a result, make more informed comparisons among these choices when building models.

This work presents preliminary results on the mapping from boundary-based classification to social welfare maximization, but it is our hope that future work will establish a bidirectional relationship such that insights in welfare maximization may be translated for fair classification. Welfare economics as a field has a rich history of developing principles and methods of analysis centered on problems of fair representation and distribution. Linking the field with machine learning would yield complementary perspectives on fairness that would be both normative and descriptive, theoretical and implementable. 
\section*{Acknowledgements}
This work is supported in part by an NSF Graduate Research Fellowship and NSF Grant \#CCF-1718549.





\bibliography{example_paper}
\bibliographystyle{icml2018}

\end{document}